\documentclass[10pt,twocolumn]{article}
\usepackage{lmodern, upquote}
\usepackage{geometry}
\usepackage{titlesec}

\titlespacing{\section}{0em}{*1}{*1}[0em]
\titlespacing{\subsection}{0em}{*1}{*0.7}[0em]
\titlespacing{\subsubsection}{0em}{*0.8}{*0.5}[0em]
\titlespacing{\paragraph}{0em}{*0.5}{*0.5}[0em]

\newenvironment{resume}{\thispagestyle{empty}\subsection*{Résumé}\em}{}
\newenvironment{motscles}{\subsection*{Mots-clés}\em}{}
\renewenvironment{abstract}{\subsection*{Abstract}\em}{}
\newenvironment{keywords}{\subsection*{Keywords}\em}{}

\usepackage{scalerel}

\def\msquare{\mathord{\scalerel*{\Box}{gX}}}

\usepackage{graphicx}
\graphicspath{{figures/}}
\usepackage[table,svgnames,x11names,dvipsnames]{xcolor}
\usepackage[T1,hyphens]{url}
\usepackage[hidelinks,urlcolor=blue,colorlinks=true,citecolor=blue]{hyperref}
\usepackage{amsmath, amsfonts, amssymb}
\usepackage{multirow, booktabs, cleveref}
\usepackage{soul}

\usepackage[most]{tcolorbox}
\newtcbox{\inlinebox}[1][]{enhanced, box align=base, nobeforeafter, colback=gray, colframe=gray,
 fontupper=\bfseries\color{white}, left=0pt, right=0pt, top=0pt, bottom=0pt, boxsep=0pt, #1} 

\title{Improving Patent Mining and Relevance Classification using Transformers\thanks{6th National Conference on Practical Applications of Artificial Intelligence, 2021, Bordeaux, France}} 
\author{Théo Ding\textsuperscript{1,2}, Walter Vermeiren\textsuperscript{2}, Sylvie Ranwez\textsuperscript{1}, Binbin Xu\textsuperscript{1}
\\[6pt]
\textsuperscript{1} EuroMov Digital Health in Motion, Univ Montpellier, IMT Mines Alès\\
\textsuperscript{2} Total Research and Technology, Seneffe, Belgium}
\date{\texttt{theo.ding@mines-ales.org, binbin.xu@mines-ales.fr}}

\begin{document}
\maketitle

\begin{resume}
L'analyse et l'exploitation de brevets sont des processus longs et coûteux pour les entreprises, mais néanmoins indispensables si elles veulent rester compétitives. Pour faire face au nombre croissant de brevets à analyser, il est possible de procéder à un filtrage préliminaire automatique, afin de n'en sélectionner qu'un nombre limité, qui est par la suite analysé par des experts. Cet article présente les résultats obtenus sur un cas d'étude industriel. Des modèles d'apprentissage profond pré-entraînés, utilisés en Traitement Automatique de la Langue, ont été fine-tunés et ré-entraînés pour améliorer la classification de brevets. La solution que nous proposons combine plusieurs traitements de l'état de l'art pour atteindre notre objectif : diminuer la charge de travail des experts tout en préservant les mesures de rappel et de précision.
\end{resume}

\begin{motscles}
Transformeurs, paramétrage, classification de brevets, traitement automatique de la langue naturelle, modèles de langage.
\end{motscles}

\begin{abstract}
{Patent analysis and mining are time-consuming and costly processes for companies, but nevertheless essential if they are willing to remain competitive. To face the overload induced by numerous patents, the idea is to automatically filter them, bringing only few to read to experts. This paper reports a successful application of fine-tuning and retraining on pre-trained deep Natural Language Processing models on patent classification. The solution that we propose combines several state-of-the-art treatments to achieve our goal$\colon$decrease the workload while preserving recall and precision metrics.}  
\end{abstract}


\section{Introduction}

Patents play one of the key roles in business intelligence. Mining state-of-the-art patents can help companies explore new or different ways of innovation to remain competitive. The conventional patent mining activity requires domain experts to manually evaluate all collected patents. A patent is considered relevant when matching the experts' current interests regarding a given subject. When the number of patents increases exponentially, the manual annotation will be too time-consuming and too costly. Tools from Natural Language Processing (NLP) have been used to filter most likely irrelevant patents for years. Unfortunately, the classification rate is quite low in many real cases. In 2018, the biggest change in NLP was the introduction of language models based on Transformer \cite{Devlin2018BERT}. BERT (Bidirectional Encoder Representations from Transformers) \cite{Devlin2018BERT} obtained state-of-the-art results on eleven natural language processing tasks, including pushing the GLUE score to 80.5\%. 
It is soon optimized by RoBERTa \cite{Liu2019RoBERTa}, which beat again the state-of-the-art results. These two have been once again improved in DeBERTa (Decoding-enhanced BERT with disentangled attention) \cite{He2020Deberta}. GPT-2, another transformer model, can not only generate coherent paragraphs of text, but also achieve state-of-the-art performances on many language modeling benchmarks \cite{Radford2019Language}. To better handle longer documents, Longformer was introduced in 2020 \cite{Beltagy2020Longformer}. 

Many successful text classification applications have been reported, including classification of complex documents like patents \cite{Choi2019Deep,Li2018DeepPatent,Roudsari2020Multi,Marco2019Patent,Hu2018Patent,Lee2020Patent}. However, patent relevance classification remains a challenging problem. The ``relevance'' itself is context dependent. One will need to retrain or fine-tune the pre-trained models for specific applications.

The main aim of the project is to study the feasibility of building an automatic patent relevance evaluation tool to pre-classify collected patents and reduce the annotation load from experts for further evaluations. All patents classified as relevant by this tool, whether they are truly relevant or not, will then be manually appraised by experts. False positives are eventually rejected during the experts' manual inspection.
One critical aspect in patent relevance classification is: misclassifying as few relevant patents as possible is vital as each one holds a potential threat against the business, even though this leads to a slightly larger number of patents left to be manually labeled.

In this work, we report a successful application of fine-tuning on pre-trained deep NLP models on patent relevance classification, by benchmarking on state-of-the-art transformer models in order to find the most appropriate model(s) for this application. 

\section{Data and Models}

\subsection{Data}

As aforementioned, the data that we consider are patents in English and, more precisely, their titles combined with the claims. Indeed, claims are the part containing key information about patents and highlight their essential elements \cite{Zhang2015Patent}. Titles contain more general information, so they could also be used as features.

\begin{figure}[!h]
\begin{tcolorbox}
\small
1. A computer-implemented method for processing a 4D seismic signal relative to a subsoil, the subsoil including a zone subject to extraction and/or injection, the method comprising: providing the 4D seismic signal; identifying a part of the 4D seismic signal corresponding to a zone of the subsoil distinct from the zone subject to extraction and/or injection; determining a noise model of the 4D seismic signal based on the identified part of the 4D seismic signal; and processing the 4D seismic signal based on the noise model.
\end{tcolorbox}
\caption{One claim from patent ``Processing a 4D Seismic Signal Based on Noise Model'' \href{https://patentscope.wipo.int/search/en/detail.jsf?docId=WO2019053484}{\sc{WO2019053484}}  \cite{Hubans2017Processing}}
\label{fig:patent}
\end{figure}

Every year, thousands of patents for one topic only are being evaluated by Total SE. Amongst these patents, in general, only $10\%$ are considered relevant. As imbalanced classes are problematic for machine learning model training, we used an oversampling technique to solve this issue: by randomly duplicating relevant patents until their number is equal to the one of the other class. 

\subsubsection{Dataset topics}

All the datasets in this work have been manually evaluated by experts. Two confidential topics have been studied here. The first topic (Topic A) has been followed by Total SE for more than 20 years, so relevant patents can be easily identified by the experts. This dataset is thus more homogeneous in terms of text diversity. It includes data from two successive years (named A1 and A2) as well as data from the first three quarters of a third year (named AT1, AT2 and AT3). The second topic (Topic B) has been tracked for only a few years and is a more heterogeneous group than topic A. It is composed of three sequential datasets (named B1, B2 and B3) over a period of one and a half years.

Topic A and topic B were only used to evaluate the transformers' performances on different topics. Better results with topic A than with topic B are to be expected due to the nature of the datasets themselves and do not represent the point of interest of this study. Topic B has been studied to know whether newer topics could also benefit from the classification automation.

Any reference to a combination of several datasets of the same topic will be presented by concatenating the corresponding dataset names if no confusion may occur.

In the following table, size refers to the number of patents in each dataset.

\begin{table}[!htb]
  \caption{Approximate number of patents for each dataset.}
  \label{tab:data_assets}%
  \vspace{0.5em}
  \centering
    \begin{tabular}{|c|c|c|c|}
    \hline
    Topic A & Size & Topic B & Size\\
    \hline
    A1 & 3,500 &  B1 & 6,500 \\
    A2 & 4,500 &  B2 & 1,500 \\
    AT1 & 1,500 &  B3 & 3,500 \\
    AT2 & 1,500  & & \\
    AT3 & 1,500  & & \\
    \hline
    \end{tabular}%
\end{table}%

\subsubsection{Training and testing strategies} \label{Training and testing strategies}

The classification strategy implies to rely partly or entirely on history datasets for model training to predict the relevance of new datasets. 
For the datasets in Table \ref{tab:data_assets}, this led to the following train / test configurations$\colon$
\begin{table}[!htb]
  \caption{Training and testing dataset configurations.}
  \label{tab:trainTestComb}%
\vspace{0.5em}
  \centering
    \begin{tabular}{|l|l|}
    \toprule
    Training datasets & Testing datasets \\
    \midrule
    A1 & A2 \\
    A12 & AT123 \\
    AT12 & AT3 \\
    A12T12 & AT3 \\
    B1 & B2 \\
    B1 & B3 \\
    B12 & B3 \\
    \bottomrule
    \end{tabular}%
\end{table}%

By default, for every model, the maximum sequence length is set to 512 tokens (except for Longformer where it has been set to 1,024 tokens), the batch size to 64, the learning rate to $2e-5$ with a linear scheduler with no {warmup} steps. Each model has been trained for 20 epochs.

Several pre-processings have been performed on these datasets, although not all of them have been used for each experiment. The datasets have firstly been re-sampled to avoid the imbalanced-class problem. For a given patent, its title and claims have been concatenated into a single string. The maximum sequence length is set to 512 tokens (about 450 words). However, when considering all the claims concatenated for one patent in our dataset, the median sequence length is closer to 1,000 tokens. In consequence, each patent's group of claims longer than 512 tokens have been truncated into chunks up to a maximum of 512 tokens. Then, each chunk has been processed by the transformers as one separate patent with the same label as the other chunks. For a given patent, the mean and the median of the probabilities of each chunk, returned by the transformers, have been calculated, and the resulting probability has been assigned to that patent.

To indicate which strategy has been used for which experiment in Tables \ref{1st_results}, \ref{2nd_results}, \ref{3rd_results} and \ref{tab:Extensive BERT}, the following naming rules have been used$\colon$
\begin{itemize}
    \item Training and testing datasets are separated by an underscore.
    \item Suffix \textbf{R}$\colon$dataset has been re-sampled, its name is followed by a suffix \textbf{R}.
    \item Suffix \textbf{T}$\colon$titles have been concatenated to claims.
    \item Suffix \textbf{EX}$\colon$dataset has been extended -- longer patents truncated into shorter chunks of 512 tokens.
\end{itemize}
If the training dataset has the title and/or the extended modifications, the test dataset will have the same tweaks.

For example, \textbf{A1-TR-EX\_A2-T-EX} indicates that the training dataset A1 (first dataset of topic A) has the titles of the patents concatenated to the claims and has also been re-sampled and extended. The testing dataset, A2 (second dataset of topic A), has the title tweak as well as its claims transformed into chunks. 

In order to maximize the classification rate and evaluate the relevance of different components of patents, many combinations have been tested$\colon$experiments with and without titles, re-sampling or extending changes.

\subsection{Optimizations, Metric and Models}

To better classify the patents, the whole patent should technically be taken into account. However, the maximal patent text lengths can go beyond the transformer models' capacity.
For most transformer models used in this work, the maximum token length is 512. GPT-2 supports 1,024 tokens while Longformer can accept up to 4,096 tokens.

Increasing the maximum token length requires more GPU VRAM, which amount is obviously limited. The first and simplest solution is to decrease the batch size. However, this reduction induces learning issues as smaller batches mean causing gradient to be more prone to noise. In heterogeneous datasets, patents are less similar from one batch to another, noise effect will be more important. 
Gradient accumulation has been applied in order to overcome this limitation. This technique allows accumulating the gradients over a given number of batches. After this, the model is being updated before resetting the gradient.

A possible solution is gradient checkpointing, that loads part of the activation network and recomputes the missing activation before operating the back-propagation, rather than loading the full graph. It trades off time for memory. However, as communications inter-GPUs take considerable time, this technique is less interesting for multi-GPU configurations. 

Another way to limit the GPU VRAM consumption is to reduce the vocabulary size and token interaction file from tokenizers. The idea is to build the vocabulary only from the current dataset. For all the benchmarked models except DeBERTa, a unique list of all the tokens and interactions between those tokens encountered in each training and test set couples are kept. This helped to reduce by around 50\% the vocabulary size and by 20\% the interaction file for all the considered models for topic A, and by 60\% and 25\% for the same files for topic B. Reducing the vocabulary directly impacts the size of the input tensor fed to the models and thereby the amount of VRAM needed, while not increasing the execution time.

\subsection{Introduction to specific metric}

Patent relevance evaluation is not a simple text classification problem. Reducing the workload from experts' manual annotations is also crucial. Common metrics only, like recall, precision or even $F1$ score, were not enough. A new metric is proposed here, taking into account the percentage of patents that will not be manually processed, and the proportion of relevant patents properly classified$\colon$
\begin{equation}
\mathcal{M} = (4 * \mathrm{recall} + \mathrm{precision} + \mathrm{patents\ left} )\ /\ 6 
\end{equation}
The \emph{patents left} changes according to the number of irrelevant patents incorrectly classified as relevant, increasing when this number decreases, and vice-versa. It represents the amount of unnecessary workload that needs to be manually done.
The coefficient $4$ multiplying the recall is to rank models according to their ability to answer the initial problem$\colon$high recall, low number of patents left to be manually handled. Its value has been set to follow the experts' view on the models' performances: by asking the experts which model would be preferred over which other according to their recall, precision and patents left scores.

As a reminder, in the business intelligence context, having a few more patents to be manually labeled (so more false positives) than missing truly relevant patents (false negatives) will always be preferred.

\subsection{Models}

The transformer models used in our study are state-of-the-art models$\colon$BERT, GPT2, RoBERTa, DeBERTa and Longformer. 

Since BERT \cite{Devlin2018BERT} is the first deeply bidirectional, unsupervised language representation (Transformer architecture), it is used here as a baseline model for all the different modifications in the experiments.
GPT2 \cite {Radford2019Language} is the first large scale generative model capable of generating realistic texts. Its token prediction ability can also be used for classification tasks when using specific tokens as class labels \cite{Xu2020Pre-Training}. 
RoBERTa \cite{Liu2019RoBERTa} is known to perform better in general compared to BERT, partly thanks to its rigorous training with ten times more data than BERT. 
DeBERTa was part of the combination of models giving the second-best results on GLUE and SuperGLUE benchmarks and, as both gather a wide range of general Natural Language Understanding tasks, implementing this model was promising.

Finally, as most of these models can only work with sequences of up to 512 tokens (1,024 for GPT2), and as our dataset has a median length of around 1,000 tokens for the considered patents, trying to implement Longformer with its maximum sequence length of 4,096 was obviously an interesting track. Moreover, Longformer is known to perform better than RoBERTa on tasks involving long documents.

All of the transformers have been adapted from Hugging Face \cite{Wolf2020Transformers}.

\section{Results}

{To evaluate the effects of each pre-processing on the datasets as well as the choices of the datasets themselves, 12 experiments have been performed. Furthermore, more combinations, presented hereinbelow, have been tested using BERT only.}

\subsection{Extensive BERT comparison}

\subsubsection{BERT results table introduction} \label{BERT results table introduction}

To have a baseline for as many experiments as possible and to compare them to one another, running BERT in its base uncased version appeared as the simplest and most obvious solution. The experiment naming rules are those aforementioned in Section \ref{Training and testing strategies}. The models' name column has been replaced by the specific features, if any, as a unique model only has been used.

\begin{table}[!htb]
  \caption{Extensive BERT experiments. From left to right, the columns are$\colon$Specific feature, Recall, Precision, Percentage, Score $\mathcal{M}$, Score $F1$, Maximal Memory usage in GB and Total computing time in minutes.}
  \label{tab:Extensive BERT}%
  \vspace{1em}
\setlength{\tabcolsep}{2.8pt}
\small
  \centering
    \begin{tabular}{|l|c|c|c|c|c|c|c|}
    \toprule
    Feature & R     & P     & \multicolumn{1}{l|}{\%} & $\mathcal{M}$     & F1    & \multicolumn{1}{l|}{GB} & \multicolumn{1}{l|}{Time} \\
    \midrule
    \multicolumn{8}{|l|}{\inlinebox{(1)} \hfill  A1\_A2} \\
    \midrule
          & 0.8330 & 0.6618 & 12.7  & 0.8252 & 0.7376 & 16    & 60 \\
    \midrule
    \multicolumn{8}{|l|}{\inlinebox{(2)} \hfill  A1-R\_A2} \\
    \midrule
    NV\textsubscript{Mem} & 0.9176 & 0.6131 & 15.1  & 0.8709 & 0.7351 & 18.8  & 82 \\
            & 0.9039 & 0.5783 & 15.7  & 0.8546 & 0.7054 & 16    & 79 \\
    Mem     & 0.9039 & 0.5783 & 15.7  & 0.8546 & 0.7054 & 18.8  & 80 \\
    Batch-8 & 0.9497 & 0.4814 & 19.9  & 0.8629 & 0.6390 & 18.8  & 153 \\
    \midrule
    \multicolumn{8}{|l|}{\inlinebox{(3)} \hfill  A1-R-EX\_A2-EX} \\
    \midrule
    $\mathrm {BERT_{\bar{X}}}$   & 0.9542 & 0.4691 & 20.5  & 0.8629 & 0.6290 & 16.5  & 106 \\
    $\mathrm {BERT_{\tilde{X}}}$ & 0.9542 & 0.4691 & 20.5  & 0.8629 & 0.6290 & 16.5  & 106 \\
    \midrule
    \multicolumn{8}{|l|}{\inlinebox{(4)} \hfill  A1-T\_A2-T} \\
    \midrule
          & 0.8673 & 0.6413 & 13.6  & 0.8436 & 0.7374 & NA     & 58 \\
    \midrule
    \multicolumn{8}{|l|}{\inlinebox{(5)} \hfill  A1-TR\_A2-T} \\
    \midrule
    NV    & 0.9497 & 0.5912 & 16.2  & 0.8873 & 0.7287 & 17.6  & 79 \\
          & 0.9474 & 0.5550 & 17.2  & 0.8780 & 0.6999 & 34    & 78 \\
    1-GPU & 0.9451 & 0.5485 & 17.3  & 0.8751 & 0.6941 & 25.9  & 155 \\
    \midrule
    \multicolumn{8}{|l|}{\inlinebox{(6)} \hfill  A1-TR-EX\_A2-T-EX} \\
    \midrule
    $\mathrm {BERT_{\bar{X}}}$   & 0.8970 & 0.6164 & 14.6  & 0.8580 & 0.7307 & 16.6  & 110 \\
    $\mathrm {BERT_{\tilde{X}}}$ & 0.8970 & 0.6164 & 14.6  & 0.8580 & 0.7307 & 16.6  & 110 \\
    \midrule
    \multicolumn{8}{|l|}{\inlinebox{(7)} \hfill  A12-TR\_AT123-T} \\
    \midrule
          & 0.9669 & 0.1535 & 44.3  & 0.7743 & 0.2650 & NA     & 142 \\
    \midrule
    \multicolumn{8}{|l|}{\inlinebox{(8)} \hfill  A12T12-TR\_AT3-T} \\
    \midrule
          & 0.8571 & 0.2426 & 18.2  & 0.7555 & 0.3782 & 3.2   & 173 \\
    \midrule
    \multicolumn{8}{|l|}{\inlinebox{(9)} \hfill  AT12-TR\_AT3-T} \\
    \midrule
          & 0.8312 & 0.2092 & 20.5  & 0.7287 & 0.3342 & NA     & 60 \\
    \midrule
    \multicolumn{8}{|l|}{\inlinebox{(10)} \hfill  B1-TR\_B2-T} \\
    \midrule
          & 0.6739 & 0.6392 & 17.7  & 0.7118 & 0.6561 & 16.5  & 99 \\
    \midrule
    \multicolumn{8}{|l|}{\inlinebox{(11)} \hfill  B1-R\_B3} \\
    \midrule
          & 0.7029 & 0.4462 & 23.1  & 0.6883 & 0.5459 & 16.5  & 105 \\
    \midrule
    \multicolumn{8}{|l|}{\inlinebox{(12)} \hfill  B1-TR\_B3-T} \\
    \midrule
          & 0.7695 & 0.4368 & 25.8  & 0.7282 & 0.5572 & 16.5  & 108 \\
    \midrule
    \multicolumn{8}{|l|}{\inlinebox{(13)} \hfill  B12-TR\_B3-T} \\
    \midrule
          & 0.7200  & 0.5362 & 19.7  & 0.7208 & 0.6146 & 21.8  & 128 \\
    \midrule
    \multicolumn{8}{|l|}{\inlinebox{(14)} \hfill  B12-TR-EX\_B3-T-EX} \\
    \midrule
    $\mathrm {BERT_{\bar{X}}}$   & 0.8895 & 0.3439 & 37.9  & 0.7756 & 0.4960 & 16.6  & 186 \\
    $\mathrm {BERT_{\tilde{X}}}$ & 0.8895 & 0.3421 & 38.1  & 0.7749 & 0.4942 & 16.6  & 186 \\
    \bottomrule
    \end{tabular}%
\end{table}%

Here, \textit{NV} stands for \textit{No Vocabulary changes}; \textit{Mem} means that the memory has been more thoroughly monitored; \textit{Batch-8} is for the batch size, here only, the batch size has been decreased to 8 and the gradient accumulation step size has been increased to 8 instead of 1;\textit{$\mathrm {BERT_{\bar{X}}}$} and \textit{$\mathrm {BERT_{\tilde{X}}}$} respectively indicate that the mean or the median of probabilities, of all the chunks returned by BERT, for a given patent, is taken into account to classify it; \textit{1-GPU} means that the model is run on single-GPU mode, without any parallelization (instead of 4 for the other tests).

All results going forward will, by default, refer to score $\mathcal{M}$. For readability, each training-test combination has been numbered. 

\subsubsection{BERT results interpretation}

To understand these tables, here is an example with the first line of experiment \inlinebox{(2)}. The training dataset is the first one of topic A (A1), which has been re-sampled, the testing dataset is the second dataset of topic A (A2). Line $1$ has BERT's vocabulary file not altered, and the memory has been thoroughly monitored during the whole run. The model's recall score is $0.9176$ for a precision of $0.6131$. The percentage of patents left to be classified manually is $15.1\%$ of the total number of patents. This means that, if we decided to take into account the decision of the model for each patent, $15.1\%$ of all the patents would have been considered relevant by the model and manually classified, the rest would not have been further processed. The $\mathcal{M}$ score for this model is $0.8709$ and its $F1$ score is $0.7351$. The maximal amount of VRAM consumed is $18.8$GB and the total execution time is 82 minutes.

Experiment \inlinebox{(1)} and line $2$ of experiment \inlinebox{(2)} show that re-sampling the data yields better results (sequentially, Score $\mathcal{M}$=0.8252 against 0.8546, 2.94\% better).

Experiments \inlinebox{(1)} and \inlinebox{(4)} show that concatenating the titles to the claims yields better results (sequentially, Score $\mathcal{M}$=0.8252 against 0.8436, 1.84\% better).

Experiment \inlinebox{(1)} and line $2$ of experiment \inlinebox{(5)} show that combining re-sampling with concatenation of titles yield better results (sequentially 0.8252 against 0.8780, 5.28\% better). Similar results can be observed with topic B.

Lines $1$ and $2$ of experiment \inlinebox{(5)} show that using the original vocabulary from the tokenizer yields better results (sequentially 0.8873 against 0.8780, 0.93\% better). The execution time is similar.

Lines $2$ and $3$ of experiment \inlinebox{(2)} allow to check the influence of noise on the results as all randomness may not have been properly seeded. Here, the identical results confirmed that noise was not an issue.

Lines $2$ and $4$ of experiment \inlinebox{(2)} show that decreasing the batch size to 8 while using a gradient accumulation with a step size of 8 yields better results (sequentially 0.8546 against 0.8629, 0.83\% better).

Experiments \inlinebox{(3)}, \inlinebox{(6)} and \inlinebox{(14)} show that $\mathrm {BERT_{\bar{X}}}$ and $\mathrm {BERT_{\tilde{X}}}$ yield almost identical results. However, experiments \inlinebox{(2)} and \inlinebox{(3)}, experiments \inlinebox{(5)} and \inlinebox{(6)}, and experiments \inlinebox{(13)} and \inlinebox{(14)} show that extending datasets does not systematically yield better results (respectively 8.3\% better, 2.0\% worse, 5.41\% better with $\mathrm {BERT_{\bar{X}}}$ or 5.48\% better with $\mathrm {BERT_{\tilde{X}}}$).

Lines $12$ and $3$ of experiment \inlinebox{(5)} show that training and testing on only 1 GPU compared to 4 does not change significantly the results (0.29\%), but the computing time has been drastically increased, here by two folds.

Experiments \inlinebox{(8)} and \inlinebox{(9)} show that using A12 along with AT12, for training, only improved the results by 2.7\% but increased the total computing time by three folds (60 minutes when training on AT12, against 173). Depending on the retraining frequency, it might be more interesting to only use the most recent data as train dataset.

Experiments \inlinebox{(12)} and \inlinebox{(13)} show that adding B2 to B1 as training dataset and predicting on B3 did not yield way better results than using only B1 as training dataset (sequentially 0.7282 against 0.7208, so 0.74\% better).

Similar results can be observed for topic B and its dataset (experiments \inlinebox{(10)} through \inlinebox{(14)}). The rest of the results' interpretation will be left to the reader.

The memory usage is the maximal amount used in the whole run. Values are just indicative and may be erroneous; models may work properly with lower amounts of memory as explained in \Cref{GPU Memory usage and training time}.

\subsubsection{BERT results conclusion}

To conclude on this table, using the titles and re-sampling tweaks consistently led to better results (5.28\% better for topic A). For two out of three cases, extending the datasets have yielded better results, even up to 5\% for topic B. This may be an interesting track to follow for future studies. Moreover, as memory scarcity is an issue, decreasing the batch size, as well as limiting the vocabulary files, are promising trails that seem to not or only slightly decrease the performances. Besides, using the original vocabulary files yields better results and noise is not an issue for experiment reproducibility when seeds have been properly set.

\subsection{Full results}

\subsubsection{Full results table introduction} \label{Full results table introduction}

Thanks to the powerful computing resources made available by Total SE, it has been possible to massively train different models on different dataset combinations with different strategies. 

Hereunder, the detailed metrics for all the experiments are presented in Tables \ref{1st_results}, \ref{2nd_results} and \ref{3rd_results}.
Only the best results, according to score $\mathcal{M}$ amongst the 20 epochs, have been kept and sorted in the result tables in decreasing order.

\begin{table}
\setlength{\tabcolsep}{4pt}
  \caption{Experiments 1 through 5. (Results ranked by descending metric $\mathcal{M}$ order).}
  \label{1st_results}
  \vspace{1em}
  \small
  \resizebox{\columnwidth}{!}{%
    \begin{tabular}{|l|c|c|c|c|c|}
    \toprule
    MODEL & \multicolumn{1}{|c|}{R }  &  \multicolumn{1}{|c|}{P}  &  \multicolumn{1}{|c|}{ \% } & \multicolumn{1}{|c|}{$\mathcal{M}$} & \multicolumn{1}{|c|}{F1}  \\
    \midrule
    \multicolumn{6}{|l|}{\inlinebox{(1)} \hfill  A1-R\_A2}  \\
    \midrule
    {\sc Mean}  & 0.9359 & 0.5603 & 16.8  & \textbf{0.8717} & 0.7009 \\
    DeBERTa & 0.9428 & 0.5372 & 17.7  & 0.8711 & 0.6844 \\
    {\sc Median} & 0.9359 & 0.5542 & 17.0  & 0.8703 & 0.6962 \\
    BERT  & 0.9039 & 0.5783 & 15.7  & 0.8546 & \textbf{0.7054} \\
    Longformer & \textbf{0.9565} & 0.4274 & 22.5  & 0.8541 & 0.5908 \\
    GPT2  & 0.9291 & 0.4975 & 18.8  & 0.8532 & 0.6480 \\
    RoBERTa & 0.8787 & \textbf{0.5818} & \textbf{15.2}  & 0.8388 & 0.7001 \\
    \midrule
    \multicolumn{6}{|l|}{\inlinebox{(2)} \hfill  A1-R-EX\_A2-EX} \\
    \midrule
    $\mathrm {DeBERTa_{\bar{X}}}$ & 0.9519 & 0.4929 & 19.4  & \textbf{0.8670} & 0.6495 \\
    $\mathrm {DeBERTa_{\tilde{X}}}$ & 0.9519 & 0.4917 & 19.5  & 0.8667 & 0.6485 \\
    $\mathrm {BERT_{\bar{X}}}$ & \textbf{0.9542} & 0.4691 & 20.5  & 0.8629 & 0.6290 \\
    $\mathrm {BERT_{\tilde{X}}}$ & \textbf{0.9542} & 0.4691 & 20.5  & 0.8629 & 0.6290 \\
    {\sc Mean}  & 0.8856 & 0.6627 & 13.5  & 0.8600  & 0.7581 \\
    {\sc Median} & 0.8833 & \textbf{0.6644} & \textbf{13.4}  & 0.8588 & \textbf{0.7583} \\
    $\mathrm {GPT2_{\bar{X}}}$ & 0.9176 & 0.5434 & 17.0  & 0.8560 & 0.6826 \\
    $\mathrm {GPT2_{\tilde{X}}}$ & 0.9176 & 0.5434 & 17.0  & 0.8560 & 0.6826 \\
    $\mathrm {RoBERTa_{\bar{X}}}$ & 0.9314 & 0.4933 & 19.0  & 0.8538 & 0.6450 \\
    $\mathrm {RoBERTa_{\tilde{X}}}$ & 0.9314 & 0.4933 & 19.0  & 0.8538 & 0.6450 \\
    \midrule
    \multicolumn{6}{|l|}{\inlinebox{(3)} \hfill  A1-TR\_A2-T} \\
    \midrule
    {\sc Median} & 0.9451 & 0.5833 & 16.3  & \textbf{0.8826} & 0.7214 \\
    {\sc Mean}  & 0.9405 & 0.5855 & 16.2  & 0.8801 & 0.7217 \\
    BERT  & 0.9474 & 0.5550 & 17.2  & 0.8780 & 0.6999 \\
    BERT-L & 0.9245 & \textbf{0.6057} & \textbf{15.4}  & 0.8739 & \textbf{0.7319} \\
    DeBERTa & 0.9451 & 0.5329 & 17.8  & 0.8717 & 0.6815 \\
    RoBERTa-L & \textbf{0.9588} & 0.4540 & 21.3  & 0.8622 & 0.6162 \\
    Longformer & 0.9268 & 0.5407 & 17.3  & 0.8614 & 0.6830 \\
    GPT2-M & 0.9062 & 0.5928 & \textbf{15.4}  & 0.8592 & 0.7167 \\
    RoBERTa & 0.9085 & 0.5712 & 16.0  & 0.8561 & 0.7014 \\
    GPT2  & 0.9176 & 0.5161 & 17.9  & 0.8500  & 0.6606 \\
    \midrule
    \multicolumn{6}{|l|}{\inlinebox{(4)} \hfill  A1-TR-EX\_A2-T-EX} \\
    \midrule
    $\mathrm {DeBERTa_{\bar{X}}}$ & \textbf{0.9588} & 0.5244 & 18.4  & \textbf{0.8787} & 0.6780 \\
    $\mathrm {DeBERTa_{\tilde{X}}}$ & \textbf{0.9588} & 0.5224 & 18.5  & 0.8782 & 0.6764 \\
    $\mathrm {RoBERTa_{\bar{X}}}$ & 0.9474 & 0.5049 & 18.9  & 0.8668 & 0.6587 \\
    $\mathrm {RoBERTa_{\tilde{X}}}$ & 0.9474 & 0.5049 & 18.9  & 0.8668 & 0.6587 \\
    {\sc Mean}  & 0.8810 & 0.6912 & \textbf{12.8}  & 0.8626 & \textbf{0.7746} \\
    {\sc Median} & 0.8787 & \textbf{0.6919} & \textbf{12.8} & 0.8612 & 0.7742 \\
    $\mathrm {BERT_{\bar{X}}}$ & 0.8970 & 0.6164 & 14.6  & 0.8580 & 0.7307 \\
    $\mathrm {BERT_{\tilde{X}}}$ & 0.8970 & 0.6164 & 14.6  & 0.8580 & 0.7307 \\
    $\mathrm {GPT2_{\bar{X}}}$ & 0.9039 & 0.5758 & 15.8  & 0.8541 & 0.7035 \\
    $\mathrm {GPT2_{\tilde{X}}}$ & 0.9039 & 0.5750 & 15.8  & 0.8539 & 0.7028 \\
    \midrule
    \multicolumn{6}{|l|}{\inlinebox{(5)} \hfill  A12T12-TR\_AT3-T} \\
    \midrule
    BERT-L & \textbf{0.9221} & 0.1578 & 30.1  & \textbf{0.7654} & 0.2694 \\
    BERT  & 0.8571 & 0.2426 & 18.2  & 0.7555 & 0.3782 \\
    DeBERTa & 0.8701 & 0.2030 & 22.1  & 0.7512 & 0.3292 \\
    Longformer & 0.8571 & 0.2082 & 21.2  & 0.7448 & 0.3350 \\
    RoBERTa & 0.8701 & 0.1791 & 25.0  & 0.7424 & 0.2971 \\
    {\sc Median} & 0.8312 & 0.2278 & 18.8  & 0.7346 & 0.3575 \\
    GPT2  & 0.8312 & 0.2169 & 19.7  & 0.7312 & 0.3441 \\
    {\sc Mean}  & 0.8182 & 0.2308 & 18.3  & 0.7272 & 0.3600 \\
    RoBERTa-L & 0.7792 & 0.2521 & 15.9  & 0.7083 & \textbf{0.3810} \\
    GPT2-M & 0.6883 & \textbf{0.2611} & \textbf{13.6} & 0.6523 & 0.3786 \\
    \bottomrule
    \end{tabular}%
}
\end{table}%

In the first row of the table, the \textit{R} stands for Recall, the \textit{P} is Precision, the \textit{\%} is the percentage of patents classified as relevant, whether there gold label is relevant, and so left to be manually classified, the \textit{$\mathcal{M}$} refers to the proposed score, and the \textit{F1} is the $F1$ score. The next row has the training and testing datasets used and all the dataset modifications that have been applied. Then, each row represents the performances of a model on this training-test combination.

{\sc Mean} is the mean of probabilities given by all the models for each patent; {\sc Median} is their median. The suffixes $\msquare_{\bar{X}}$ or $\msquare_{\tilde{X}}$ of certain models follow the same nomenclature as in \Cref{BERT results table introduction}. Whenever the large version of a transformer was involved, a suffix \textit{L} has been added to the model's name. For GPT2, the \textit{M} stands for the medium-sized version. Finally, the maximum sequence length used has been set to 512 for every model except for Longformer where it was 1,024 tokens.

\subsubsection{Full results interpretation}

Similarly to the previous \Cref{tab:Extensive BERT}, for each experiment, each line corresponds to one model with its values for recall, precision, patents left, score $\mathcal{M}$ and score $F1$.

In 10/12 experiments, DeBERTa is in the top two best models, followed by BERT/BERT large with 8/12. Longformer is in 3/9 top twos, when RoBERTa is in 2/12, and GPT2 reached it only once out of twelve experiments. The {\sc Mean} and {\sc Median} of all probabilities performed better than individual models respectively only three times and twice out of twelve. Note that these {\sc Mean} and {\sc Median} came from the best models obtained within the 20 epochs, so they may not be the best combinations that could exist in theory. However, testing all the possible cases is not possible. As an example, there are $2 * (21^8 - 1) \approx 7.6e9$ possible combinations for experiment \inlinebox{(9)} only.

In 8 experiments out of the 12 ones, {\sc Mean} and {\sc Median} are amongst the top three models. In 6/12 experiments, the {\sc Mean} is better than the {\sc Median} of probabilities. However, the mean difference in score $\mathcal{M}$ when the {\sc Mean} is above the {\sc Median} is only 0.0071, whereas this mean difference is 0.0238 when the {\sc Median} is above the {\sc Mean}. This shows that, although the {\sc Mean} is above the {\sc Median} in the same number of experiments, the {\sc Median} should be the first choice as voting strategy.

To interpret part of the results in a clearer way, two experiments, designated by their numbers, will be compared to each other. To do so, the mean of score $\mathcal{M}$ of each individual model will be calculated for each couple of experiments compared. In other words, this mean will not include neither {\sc Mean} nor {\sc Median}. In the form of a listing, the first and the second experiments will be joined by an "\&" sign, followed by a coma and two scores, referring sequentially to the first and the second experiments, and joined by an "\&". Finally, a colon "$\colon$" will lead the interpretations.

Interpretation results$\colon$
\begin{itemize}
    \item \inlinebox{(1)} \& \inlinebox{(2)}, 0.8544 \& 0.8599$\colon$extending the datasets yields better results (0.55\%).
    \item \inlinebox{(1)} \& \inlinebox{(3)}, 0.8544 \& 0.8641$\colon$concatenating the titles to the claims yields better results (0.97\%).
    \item \inlinebox{(3)} \& \inlinebox{(4)}, 0.8641 \& 0.8643$\colon$extending the datasets yields very slightly better results (0.02\%).
    \item \inlinebox{(5)} \& \inlinebox{(6)}, 0.7314 \& 0.7051$\colon$although A1 and A2 are not part of the same year as AT123, using them alongside with AT12 for the training yields better results (2.63\%).
    \item \inlinebox{(5)} \& \inlinebox{(7)}, 0.7314 \& 0.7711$\colon$training on A12 and testing on AT123 yields better results than training on A12T12 and testing on AT3 only (3.97\%). Perhaps the patents in A12 are more similar to those in AT12 than in AT3. AT3 has also been published later than AT12 compared to A12.
    \item \inlinebox{(8)} \& \inlinebox{(12)}, 0.7398 \& 0.7020$\colon$training on B1 and testing on B2 yields better results than training on B1 and testing on B3 (3.78\%). Once again, this may be due to the chronological distance being longer between B1 and B3 than between B1 and B2. Topic B being broader, patents considered to be relevant may vary faster with time than with other topics. This can mean that having recent datasets for training may be more critical than having more datasets but less updated.
    \item \inlinebox{(9)} \& \inlinebox{(12)}, 0.7364 \& 0.7020$\colon$training on B12 and testing on B3 yields better results than training on B1 and testing on B3 (3.44\%). The possible interpretation is the same as the previous ones.
    \item \inlinebox{(11)} \& \inlinebox{(12)}, 0.6717 \& 0.7020$\colon$concatenating the titles to the claims yields better results (3.03\%).
    \item \inlinebox{(9)} \& \inlinebox{(10)}, 0.7364 \& 0.7462$\colon$extending the datasets yields better results (0.98\%). Having more data involved in the decision to classify patents, in one or the other class, may yield better results with topic B, as it is broader than topic A.
\end{itemize}

\begin{table}[!htb]
\setlength{\tabcolsep}{3.5pt}
\centering
  \small
  \caption{Experiments 6 through 10. (Results ranked by descending metric $\mathcal{M}$ order).}
  \label{2nd_results}
  \vspace{1em}
    \begin{tabular}{|l|c|c|c|c|c|}
    \toprule
    MODEL & \multicolumn{1}{|c|}{R }  &  \multicolumn{1}{|c|}{P}  &  \multicolumn{1}{|c|}{ \% } & \multicolumn{1}{|c|}{$\mathcal{M}$} & \multicolumn{1}{|c|}{F1}  \\
    \midrule
    \multicolumn{6}{|l|}{\inlinebox{(6)} \hfill  AT12-TR\_AT3-T} \\
    \midrule
    BERT  & \textbf{0.8312} & 0.2092 & 20.5  & \textbf{0.7287} & 0.3342 \\
    {\sc Median} & 0.8052 & 0.2541 & 16.3  & 0.7255 & 0.3863 \\
    GPT2  & 0.7922 & 0.2699 & 15.1  & 0.7214 & 0.4026 \\
    {\sc Mean}  & 0.7922 & 0.2629 & 15.5  & 0.7196 & 0.3948 \\
    Longformer & 0.7662 & 0.2837 & 13.9  & 0.7081 & 0.4140 \\
    RoBERTa & 0.7922 & 0.2089 & 19.5  & 0.7039 & 0.3306 \\
    DeBERTa & 0.6494 & \textbf{0.4274} & \textbf{7.8}   & 0.6634 & \textbf{0.5155} \\
    \midrule
    \multicolumn{6}{|l|}{\inlinebox{(7)} \hfill  A12-TR\_AT123-T} \\
    \midrule
    {\sc Mean}  & 0.9307 & 0.2320 & 28.2  & \textbf{0.7897} & 0.3714 \\
    {\sc Median} & 0.9337 & 0.2240 & 29.3  & 0.7885 & 0.3613 \\
    BERT-L & 0.9367 & 0.2160 & 30.5  & 0.7873 & 0.3510 \\
    Longformer & 0.9578 & 0.1797 & 37.5  & 0.7839 & 0.3026 \\
    GPT2  & 0.9307 & 0.2071 & 31.6  & 0.7799 & 0.3388 \\
    BERT  & \textbf{0.9669} & 0.1535 & 44.3  & 0.7743 & 0.2650 \\
    RoBERTa-L & 0.8735 & 0.2780 & 22.1  & 0.7687 & 0.4218 \\
    DeBERTa & 0.8675 & \textbf{0.2815} & \textbf{21.7}  & 0.7660 & \textbf{0.4251} \\
    RoBERTa & 0.9518 & 0.1534 & 43.7  & 0.7652 & 0.2642 \\
    GPT2-M & 0.8404 & 0.2640 & 22.4  & 0.7435 & 0.4017 \\
    \midrule
    \multicolumn{6}{|l|}{\inlinebox{(8)} \hfill  B1-TR\_B2-T} \\
    \midrule
    DeBERTa & 0.7790 & 0.6324 & 20.7  & \textbf{0.7787} & \textbf{0.6981} \\
    Longformer & \textbf{0.8732} & 0.3970 & 37.0  & 0.7778 & 0.5459 \\
    {\sc Mean}  & 0.7862 & 0.5945 & 22.2  & 0.7749 & 0.6771 \\
    {\sc Median} & 0.8007 & 0.5525 & 24.4  & 0.7744 & 0.6538 \\
    RoBERTa & 0.7572 & 0.4988 & 25.5  & 0.7333 & 0.6014 \\
    BERT  & 0.6739 & \textbf{0.6392} & \textbf{17.7}  & 0.7118 & 0.6561 \\
    GPT2  & 0.7862 & 0.3196 & 41.4  & 0.6972 & 0.4545 \\
    \midrule
    \multicolumn{6}{|l|}{\inlinebox{(9)} \hfill  B12-TR\_B3-T} \\
    \midrule
    DeBERTa & 0.8476 & 0.4555 & 27.3  & \textbf{0.7829} & 0.5925 \\
    {\sc Mean}  & 0.8057 & 0.5146 & 22.9  & 0.7710 & 0.6281 \\
    Longformer & \textbf{0.9352} & 0.2634 & 52.0  & 0.7702 & 0.4111 \\
    {\sc Median} & 0.8114 & 0.4936 & 24.1  & 0.7696 & 0.6138 \\
    RoBERTa & 0.8876 & 0.3117 & 41.7  & 0.7625 & 0.4614 \\
    BERT-L & 0.7886 & 0.4272 & 27.0  & 0.7378 & 0.5542 \\
    BERT  & 0.7200  & 0.5362 & 19.7  & 0.7208 & 0.6146 \\
    RoBERTa-L & 0.6876 & \textbf{0.5841} & \textbf{17.2}  & 0.7105 & \textbf{0.6317} \\
    GPT2-M & 0.7562 & 0.3994 & 27.7  & 0.7096 & 0.5227 \\
    GPT2  & 0.7181 & 0.4425 & 23.8  & 0.6971 & 0.5476 \\
    \midrule
    \multicolumn{6}{|l|}{\inlinebox{(10)} \hfill  B12-TR-EX\_B3-T-EX} \\
    \midrule
    $\mathrm {BERT_{\bar{X}}}$ & \textbf{0.8895} & 0.3439 & 37.9  & \textbf{0.7756} & 0.4960 \\
    $\mathrm {BERT_{\tilde{X}}}$ & \textbf{0.8895} & 0.3421 & 38.1  & 0.7749 & 0.4942 \\
    $\mathrm {DeBERTa_{\bar{X}}}$ & 0.8343 & 0.4228 & 28.9  & 0.7655 & 0.5612 \\
    $\mathrm {DeBERTa_{\tilde{X}}}$ & 0.8343 & 0.4228 & 28.9  & 0.7655 & 0.5612 \\
    $\mathrm {GPT2_{\bar{X}}}$ & 0.7714 & 0.4540 & 24.9  & 0.7340 & 0.5716 \\
    $\mathrm {GPT2_{\tilde{X}}}$ & 0.7695 & 0.4524 & 24.9  & 0.7323 & 0.5698 \\
    $\mathrm {RoBERTa_{\bar{X}}}$ & 0.7105 & 0.5188 & 20.1  & 0.7107 & 0.5997 \\
    $\mathrm {RoBERTa_{\tilde{X}}}$ & 0.7105 & 0.5188 & 20.1  & 0.7107 & 0.5997 \\
    {\sc Median} & 0.6381 & 0.6879 & 13.6  & 0.6997 & 0.6621 \\
    {\sc Mean}  & 0.6286 & \textbf{0.7006} & \textbf{13.1}  & 0.6959 & \textbf{0.6627} \\
    \bottomrule
    \end{tabular}%
\end{table}%

In extended dataset experiments (\inlinebox{(2)}, \inlinebox{(4)} and \inlinebox{(10)}), differences in $\mathcal{M}$ scores between the same models but attributing patents' class with the median or mean of probabilities is minuscule (0.0283\% on average).

In experiment \inlinebox{(6)}, the best $F1$ score (with 0.5155) is ranked last according to score $\mathcal{M}$ as its recall is only of 0.6494, whereas the best model's recall is 0.8312. Let us remind that, strategically speaking, it is important to miss as few relevant patents as possible, so a high recall will be preferred over a high precision at equal values. In experiment \inlinebox{(7)}, the same situation occurs.

Scores with topic A are better than those of topic B. This may be due to the fact that patents in this topic are more general, and that topic B has only been tracked for a few years. These data are less refined, thus less representative, which makes the prediction task more challenging. 
Experiments \inlinebox{(3)} and \inlinebox{(9)} had similar configurations. Sequentially, without taking {\sc Mean} nor {\sc Median} into account, the maximal $\mathcal{M}$ scores are 0.8780 and 0.7829 (9.51\%) and the mean of $\mathcal{M}$ scores are 0.8641 and 0.7364 (12.77\%).

\begin{table}[!h]
\setlength{\tabcolsep}{4pt}
  \caption{Experiments 11 and 12. (Results ranked by descending metric $\mathcal{M}$ order).}
  \label{3rd_results}
  \vspace{0.5em}
\small
    \begin{tabular}{|l|c|c|c|c|c|}
    \toprule
    MODEL & \multicolumn{1}{|c|}{R }  &  \multicolumn{1}{|c|}{P}  &  \multicolumn{1}{|c|}{ \% } & \multicolumn{1}{|c|}{$\mathcal{M}$} & \multicolumn{1}{|c|}{F1}  \\
    \midrule
    \multicolumn{6}{|l|}{\inlinebox{(11)} \hfill  B1-R\_B3} \\
    \midrule
    DeBERTa & \textbf{0.8133} & 0.3509 & 33.9  & \textbf{0.7306} & 0.4902 \\
    {\sc Median} & 0.7219 & 0.4572 & 23.1  & 0.7032 & 0.5598 \\
    RoBERTa & 0.7790 & 0.3264 & 35.0  & 0.7012 & 0.4601 \\
    {\sc Mean}  & 0.7162 & 0.4619 & 22.7  & 0.7008 & \textbf{0.5616} \\
    BERT  & 0.7029 & 0.4462 & 23.1  & 0.6883 & 0.5459 \\
    Longformer & 0.6381 & 0.4085 & 22.9  & 0.6376 & 0.4981 \\
    GPT2  & 0.5543 & \textbf{0.4763} & \textbf{17.0}  & 0.6007 & 0.5123 \\
    \midrule
    \multicolumn{6}{|l|}{\inlinebox{(12)} \hfill  B1-TR\_B3-T} \\
    \midrule
    DeBERTa & \textbf{0.8057} & 0.4862 & 24.3  & \textbf{0.7641} & 0.6065 \\
    {\sc Median} & 0.7543 & 0.5083 & 21.7  & 0.7364 & 0.6074 \\
    {\sc Mean}  & 0.7505 & \textbf{0.5110} & \textbf{21.5}  & 0.7346 & \textbf{0.6080} \\
    BERT  & 0.7695 & 0.4368 & 25.8  & 0.7282 & 0.5572 \\
    Longformer & 0.7219 & 0.4019 & 26.3  & 0.6887 & 0.5163 \\
    RoBERTa & 0.6838 & 0.4793 & 20.9  & 0.6843 & 0.5636 \\
    GPT2  & 0.6438 & 0.4214 & 22.4  & 0.6445 & 0.5094 \\
    \bottomrule
    \end{tabular}%
\end{table}%

In 3 out of 4 experiments, BERT large performed better than BERT base (0.895\% better on average). In 2/4 experiments, GPT2 performed better than its medium equivalent (2.34\%). The same occurred with RoBERTa large being better than its base version (1.91\%).

Though Longformer is capable of taking much longer token sequences, it did not necessarily give better results than the other models. Longformer's large version has also been tested for only one dataset combination by decreasing the maximal sequence length to 512. However, the results were worse than those of the base version, so experiments have not been pursued. Amongst the nine experiments where Longformer has been tested, 6/9 medians of the percentage of left patents of each individual model are lower than the individual score on the same metric that Longformer has. This model may be not decisive enough for our application.

\subsubsection{Full results conclusion}

Although topic A and B have very different $\mathcal{M}$ scores, the general trend remains the same$\colon$concatenating titles to claims, re-sampling the datasets and extending them yield better results. Moreover, datasets chronologically following each other give better results as topics are better depicted in datasets closer to the dataset to be tested. This is particularly true for broad topics as the ones that have not been monitored until recently.
DeBERTa performs very consistently the best, while it was the exact contrary for GPT2. DeBERTa and BERT are more likely to be the first choices over other models when training time is limited. Even though neither {\sc Mean} nor {\sc Median} yielded the best results every time, it is still interesting to take them into account as, for most of the experiments, they were close to the top scores. {\sc Median} is usually more secure to be kept.

\subsection{GPU Memory usage and training time} \label{GPU Memory usage and training time}

\begin{figure}[!h]
\includegraphics[width=\columnwidth]{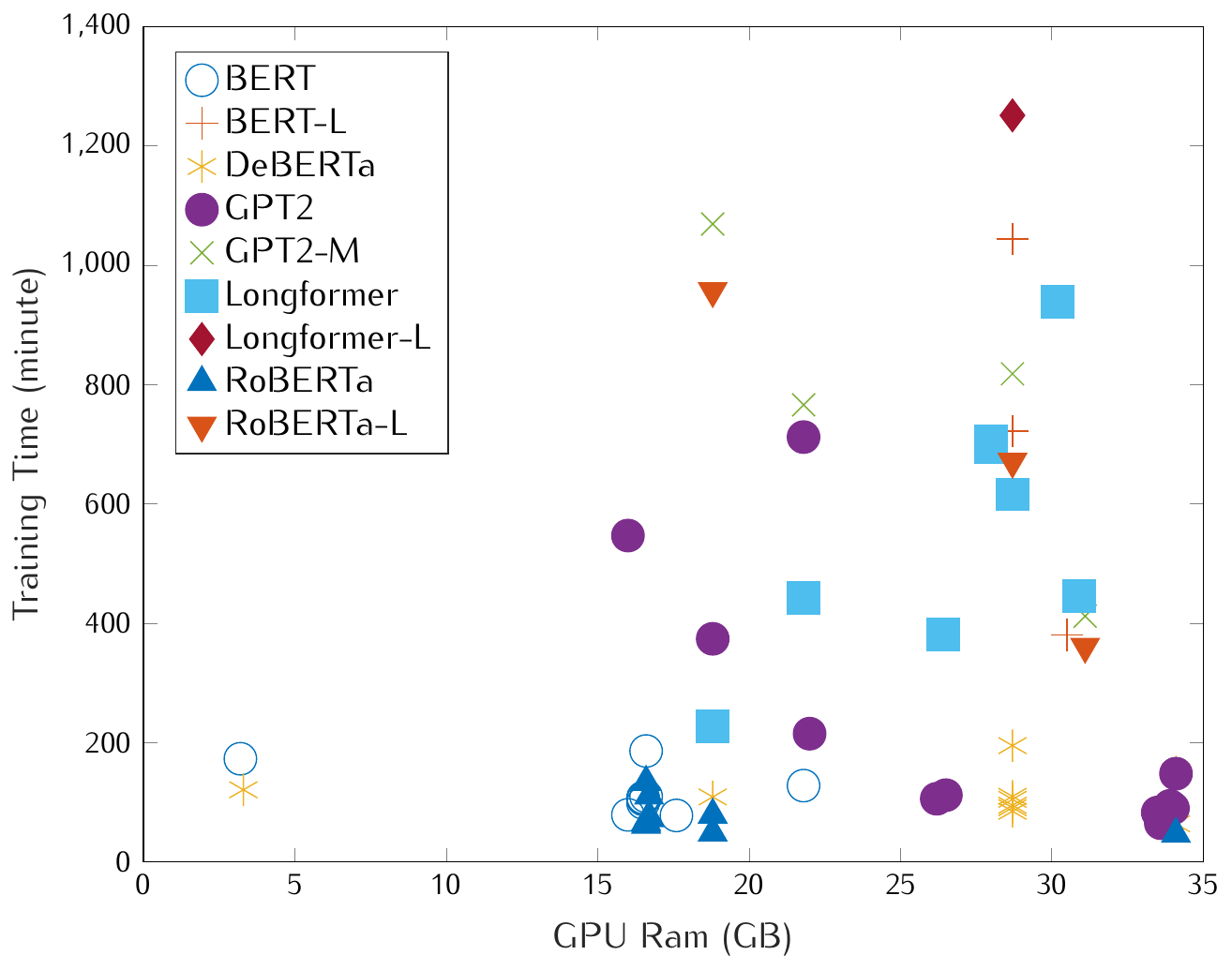}
\caption{GPU memory usages and training times for different transformer models in all the experiments.}
\label{fig_memory_and_time}
\end{figure}

The \Cref{fig_memory_and_time} shows both training time and GPU VRAM used for each model.

\subsubsection{GPU Memory usage}

\begin{figure*}[!h]
\includegraphics[width=\textwidth]{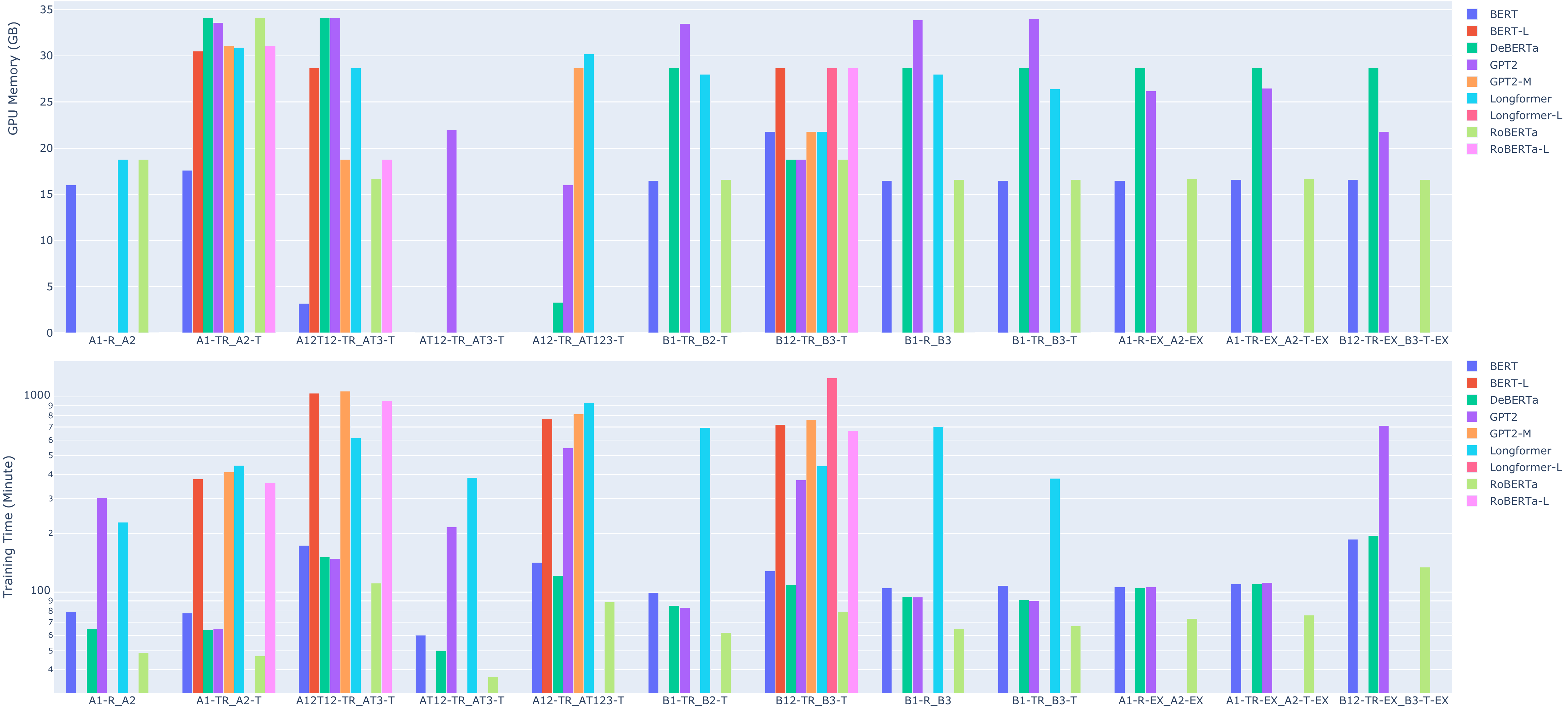}
\caption{Detailed GPU memory usage and training time for different transformer models in all the experiments. }
\label{fig_memory_time}
\end{figure*}

Total SE's HPC platform is equipped with multiple GPUs and CPUs. In this work, 4 Nvidia Tesla V100 (32GB VRAM) have been used. The peak GPU memory usage for all the tested models can be found in \Cref{fig_memory_time}. Note that GPU RAM usage reports might not be as accurate as expected. Several factors like GPU caching / RAM fragmentation / initialization etc. could prevent the program from returning the exact RAM usage information \cite{fastaiMar132021Working}. 

Using gradient accumulation allowed to run larger versions of the models, trading-off time for memory, with very limited performance influences.

For models with same data inputs and close parameter spaces, the maximal memory usage remains similar. 
These models have similar amounts of network parameters and input tensor dimensions, so the highest memory usages are at the same level. However, once the models have been initialized and are running in stable phase, much less VRAM is required. So, to correctly run these transformer models, one would need a GPU with at least 24GB or ideally 32GB of VRAM, otherwise, the model cannot even be initialized.

\subsubsection{Training time}

Memory requirements are surely an issue when it comes to training transformer models, but time can also be problematic. Here, the base models have been trained on 4 Nvidia Tesla V100 whereas larger models used only one V100 due to the gradient checkpointing technique. The preliminary benchmarking on gradient checkpointing showed, that there is no gain with multi-GPU training in the current environment. The GPU-GPU communication is much more time-consuming than the potential time-saving from the usage of multiple GPUs. Using a single V100 is faster in this case. The number of training epochs was 20 and the maximum sequence length was 512 tokens. In all tested cases, RoBERTa was the fastest, mostly followed by DeBERTa. If it was not the case, DeBERTa was usually only about 2 minutes slower than BERT or GPT2. As for GPT2, it depends on the datasets: it could perform as fast as the other models in some cases, but way slower for some other tests. 

Larger versions of the models tend to need more time to run, due to the large parameter space and the gradient checkpointing application. Longformer, whether it was in its base form or not, took way longer than the other transformers. With a sequence length of 512 tokens on B12-TR\_B3-T, base Longformer took 7 hours 20 minutes to complete the training, while large Longformer required 20 hours 50 minutes for the same task.

\section{Conclusion}

Our work aimed at determining a list of promising techniques and models to be used for business intelligence. Thanks to all the computing resources provided by Total SE, we were able to massively perform experiments with state-of-the-art transformer models and different parameter configurations. 

The exhaustive BERT experiments allowed us to evaluate the effects of different features and modifications, both on the models themselves and on the training and test datasets.

Amongst the best techniques and models we have been able to test, concatenating titles to claims, re-sampling are essential. Extending the datasets can also be very interesting but may need more work to make it a reliable solution. The best model in terms of performances and resource consumption (both VRAM and time) is DeBERTa, closely followed by BERT. The intuition that a high-performing model on the GLUE and SuperGLUE tasks would also work efficiently for our use-case has paid off. Moreover, the BERT experiments showed, that reducing the batch size, along with using gradient accumulation, gave comparable results to those of models with an unchanged batch size, and achieved this while reducing the amount of VRAM required to run.
Neither Longformer nor GPT2, both for their time consumption and performances, can be recommended for a similar task as the one in this study. 

For topic A, the best combination of models (the {\sc Median}) reached an $\mathcal{M}$ score of 0.8826 with a recall of 0.9451, a precision of 0.5833 and only 16.3\% of the total patents left for manual classification by the experts. If originally, 2,000 patents were to be processed, this model would leave only about 300 patents for manual annotation, dividing by almost ten folds the workload of the experts. However, for topic B, results are slightly worse: the best combination of models (the {\sc Mean}) reached an $\mathcal{M}$ score of 0.7710 with a recall of 0.8057, a precision of 0.5146 and 22.9\% of the total patents left for manual labeling. Results regarding topic B are worse than those of topic A, most certainly because of the broadness of subjects represented in topic B's patents.

\section{Discussion}

According to the extensive BERT experiments, it has been observed that the original vocabulary produced slightly better results than the customized vocabulary. Note that DeBERTa's vocabulary files have not been modified. Its top performances may partly be explained by this difference. Since modifying the vocabulary means the transformers will be retrained, their weights are being modified, possibly leading to a slight impact on their performances.

BERT experiments also showed better results when used gradient accumulation and reducing the batch size. However, one can argue on the veracity of this result for every model-dataset combination. Gradient accumulation should not yield better performances than using a big enough batch size.
Retraining transformer models with customized vocabularies may be usable for domain-specific tasks. However, as confirmed from the experiments, this training would require much more data to build a robust and better model. 

Experiments \inlinebox{(5)} and \inlinebox{(6)} are distinguished by the non-use of A12 as training datasets in \inlinebox{(6)}. Results are, as expected, better in \inlinebox{(5)}. However, using 11,000 examples as the training dataset against only 3,000 for a mere increase of 2.63\% in score $\mathcal{M}$ may not the most interesting time-performance trade-off. So, it is advised to use additional data only when the latter is close enough to the dataset to be predicted.
In practice, a small margin, including more patents than the group considered as relevant by the transformers, is expected, and all of them will be manually analyzed. This is to minimize the risk of missing critical patents. The margin will depend on the transformers' performances on a given topic and is to be determined accordingly.

{We are aware that not having a low $\mathcal{M}$ score is relatively expected. However, reaching a high $\mathcal{M}$ score may be a very hard achievement (90+\%), especially on broad or new topics as topic B. Although score $\mathcal{M}$, given in this study, is specific to the view of Total SE's experts on the ratio of true positives to false positives, this metric may be utilized by other companies to similarly evaluate their models' performances. For example, if missing a few more true positives but decreasing further down the workload is required, assigning a higher weight to the \emph{patents left} score and a lower one to the recall can be considered.}

The next steps would involve testing more models, larger versions, training / retraining our own transformer models, including more data from the patents like the descriptions or the abstracts, optimizing the dataset extension technique. A possible improvement might be the attribution of a weight to the probability of each chunk of claims according to their size in terms of tokens.

\renewcommand\refname{References}
\bibliographystyle{ieeetran}
\small{\bibliography{bib}}

\end{document}